# Cognitive BPM as an Equalizer: Improving Access and Efficiency for Employees with (and without) Cognitive Disabilities

Gordon Banks[1], Gates Bierhuizen[2], Katherine McCrum[3], and Ellen Wengert[4]

1,2,4 Systems Engineering Program, George Mason University, Fairfax, VA 22030 USA
3 Operations Research Program, George Mason University, Fairfax, VA 22030 USA

Corresponding author: Gordon Banks (e-mail: gbanks7@gmu.edu).

**ABSTRACT** We examine ProcessGPT, an AI model designed to automate, augment, and improve business processes, to study the challenges of managing business processes within the cognitive limitations of the human workforce, particularly individuals with cognitive disabilities. ProcessGPT provides a blueprint for designing efficient business processes that take into account human cognitive limitations. By viewing this through the lens of cognitive disabilities, we show that ProcessGPT improves process usability for individuals with and without cognitive disabilities. We also demonstrate that organizations implementing ProcessGPT-like capabilities will realize increased productivity, morale, and inclusion.

## I. INTRODUCTION

Human cognitive factors such as perception, attention, memory, language, reasoning, processing speed, and executive functions [1] have known limitations. [2][3][4][5] In contrast, Artificial intelligence (AI) systems transcend many boundaries of human cognition, particularly for memory, attention, and executive function. [6][7] Large language models such as GPT-4 already outperform humans in language areas such as lexical knowledge, grammatical sensitivity, communication ability, naming facility, and fluency. [1][8]

Exploration of AI's cognitive capabilities in the context of business process management (BPM) offers insights into knowledge-intensive processes that are predominantly driven by human activities. [9] Knowledge-intensive processes can only be partially mapped to a process model and commonly vary due to circumstances and administrative discretion. [10] Facilitation of these processes is within the domain of Cognitive BPM, which manages business processes using cognitive computing technologies. [11] We reference other researchers' AI model, ProcessGPT, as a practical representation of a Cognitive BPM solution (Fig. 1). ProcessGPT is an AI model whose goal is to suggest the best next step in a process based on a Process Knowledge Graph and extensive supporting elements. [12] While ProcessGPT is capable of process augmentation, automation, and improvement, we highlight the challenges faced by employees with cognitive disabilities by considering how ProcessGPT can augment their experience.

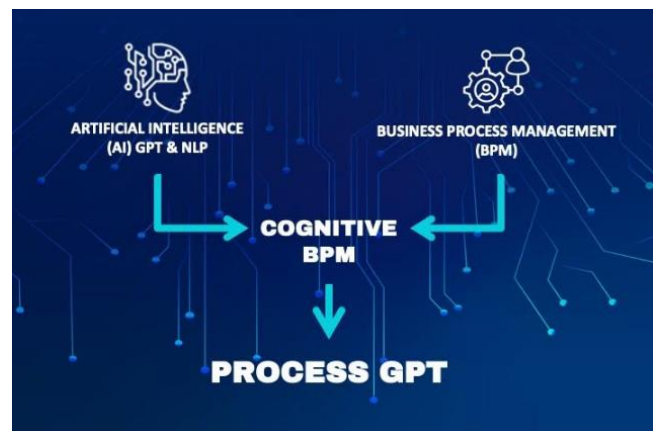

**FIGURE 1.** Relationships between AI, BPM, and ProcessGPT elements.

Through the use of AI-intensive systems, like ProcessGPT, organizations can decrease the administrative burden and cognitive load on their employees leading to increased organizational productivity (Fig. 2). ProcessGPT's potential as a facilitator for process users is profound, particularly in the context of knowledge-intensive processes, which rely heavily on human cognitive resources by their nature. Knowledge-intensive processes can prove particularly challenging for employees with cognitive disabilities, which include autism, ADHD, dyslexia, aphasia, and mild cognitive impairment. Cognitive disabilities limit the functional learning, memory, attention, and executive functions needed to perform the tasks that comprise knowledge-intensive processes. Knowing this, we establish two research questions to investigate the nexus of business processes, AI, and cognitive disabilities: 1) How can AI-intensive technology be applied to business processes to accommodate individuals with cognitive disabilities? and 2) What benefits are likely to be seen from accommodating individuals with cognitive disabilities in business processes? Through our results we aim to motivate organizations to accommodate people with cognitive disabilities in their business processes with the use of AI, which will benefit the entire organization, not just personnel with cognitive disabilities.

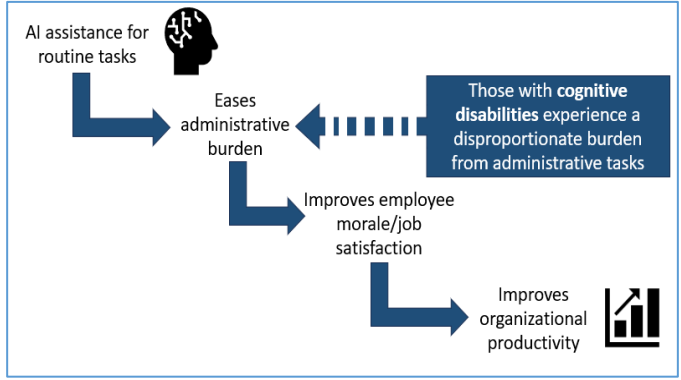

**FIGURE 2.** Mapping how AI can lead to improved organizational productivity

## II. BACKGROUND

### A. Cognitive Disabilities in the Workplace

According to the CDC, approximately 13% of the US population has some form of cognitive disability. [13] Individuals with cognitive disabilities have significant difficulties relating to:

1) learning, communication, reading, writing, or math,
2) the ability to understand or process new or complex information and learn new skills, with a reduced ability to cope independently, and / or
3) memory and attention or visual, language, or numerical thinking. [14]

These individuals face unique challenges in the workplace that depend on the cognitive resources required to properly execute their tasks. By nature, business processes shift administrative burdens to users–and their limited cognitive resources. [15] Navigating many business processes involves understanding multi-step procedures, remembering to follow up on paperwork, reading dense process documentation, or completing forms with ambiguous language and jargon. These tasks require attention to detail, adherence to procedures, and effective communication with various parties involved in the process. Doing so places demands on users' cognitive functions, reducing the time and cognitive resources they have available for other tasks.

Generally, a user's performance decreases as their cognitive load increases, and is a function of task complexity, task support, and user capabilities. User capabilities depend on cognitive factors such as attention, memory, and processing speed, in addition to non-cognitive factors such as experience, training, environmental conditions, enthusiasm, organizational culture, fatigue, and stress. [16] For any user of a business process, there exists some level of process complexity that will exceed their cognitive resources, resulting in cognitive overload. [17][15] It follows that this threshold will be lower for a user with a cognitive disability performing a task that depends on cognitive factors limited by their disability.

The World Wide Web Consortium (W3C) drafted detailed guidelines for making content usable for people with cognitive and learning disabilities. [14] These guidelines are organized by nine key objectives that designers should target:

1) Help users understand what things are and how to use them
2) Help users find what they need
3) Use clear content (text, images, and media)
4) Help users avoid mistakes
5) Help users focus
6) Ensure processes do not rely on memory
7) Provide help and support
8) Support adaptation and personalization
9) Test with real users

Each objective is decomposed into specific guidelines and design patterns. Background information is also given. For instance, within "*Objective 1: Help Users Understand What Things are and How to Use Them*," the guidelines familiarize the reader with the issues that inform this objective:

*Users with cognitive and learning disabilities may have trouble with orientation and learning. This can mean people get disoriented in a site.*

*Learning new things and remembering new information is especially difficult for people with cognitive and learning disabilities. They can also struggle or be unable to learn new design patterns. Make controls, icons and elements simple and conventional to help.*

Design patterns that support this objective are then given:

1) *Make the Purpose of Your Page Clear*
2) *Use a Familiar Hierarchy and Design*
3) *Use a Consistent Visual Design*
4) *Make Each Step Clear*
5) *Clearly Identify Controls and Their Use*
6) *Make the Relationship Clear Between Controls and the Content They Affect*
7) *Use Icons that Help the User*

Each design pattern details user needs, what to do to meet those needs, how it helps, and provides examples. Overall, the W3C guidelines emphasize using familiar design patterns with easily understood content presented in manageable chunks, as well as supporting users by anticipating cognitive shortcomings.

### B. Knowledge-Intensive Processes

BPM seeks continuous improvement of business processes, [18] and refers to poorly performing or otherwise troublesome business processes as unmanaged, [19][20] unstructured, *ad hoc*, or knowledge-intensive processes. [21][10] We adopt the *knowledge-intensive* descriptor used to initially describe ProcessGPT.

In general, business processes require users to perform administrative tasks in pursuit of an organizational goal. Knowledge-intensive business processes often lack the consistent, clear, and complete definition necessary for timely execution toward a goal. [10] As a result, users are required to either remember idiosyncrasies about a process or maintain information outside of the very systems intended to manage that process. Knowledge-intensive processes thus grow to encompass additional unmanaged elements via these mechanisms. [15] Unmanaged elements contribute to a large "digital exhaust" signature (emails, spreadsheets, documents, forms, chats, etc. that contain critical process guidelines, best practices, and policies) typical of knowledge-intensive processes. [22] This emergent complexity not only constitutes a continual drain on the cognitive resources of all workers, reducing overall productivity and morale, its unmanaged nature places the most burden on users with cognitive disabilities. Users with limited cognition will be the first to experience cognitive overload, with no recourse for accommodation.

The better users understand these processes, the better their output and, hence, their morale. The rules and procedures associated with knowledge-intensive processes are frequently associated with negative perceptions that affect both morale and output, regardless of whether those perceptions are accurate. [23] Another important factor related to employee morale and efficiency is emphasis on high-value work. Often when employees are asked about the most frustrating part of their jobs, they will mention the amount of time spent on low-value tasks that detracts from their primary work. These tasks are in many cases manual, repetitive, and can consume a lot of time. [24] Simplifying steps, eliminating confusion, and adding transparency to business processes are the key to a more efficient and satisfied workforce.

Many have already tried to address these challenges through Robotic Process Automation (RPA). RPA typically leverages commercial software to automate routine tasks, copying human actions. RPA is a growing area that has been proven to save employee time on low-value work. The Federal RPA Community of Practice estimates that as of 2021, over 1.4M hours of work have been saved due to the implementation of RPA. [25] Industry leaders such as Microsoft have been working to facilitate this type of automation by digitizing contracts. Microsoft Azure has taken physical paperwork into the cloud which opens the door for contract processes to be integrated into more efficient workflows. AI is the next logical area where contract processing should and will go. [26]

AI changes the way many industries operate. In contracting, AI can sift through an enormous amount of paperwork and understand what the content of each is. AI allows organizations an "all-knowing" power where information can be queried and analytics can be reported, all without the intervention of a human. Additionally, it can review contract content for consistency across an organization and assess risk by identifying terms that are suboptimal. Ultimately it can perform these tasks in a way that is faster and more accurate than a human. [27] This has major implications for law firms, as well as innumerable other industries.

ProcessGPT can decrease the amount of time spent on knowledge-intensive tasks by training it on a large set of business process data. The model can be further refined by user input and decisions. The goal is to automate repetitive tasks that would otherwise be conducted by human workers. This has potential to greatly improve how employees with cognitive disabilities complete their work and provide benefit to an organization.

## III. RELEVANT WORK

AI technologies already demonstrate their effectiveness in assisting individuals with cognitive disabilities, improving administrative tasks, and ensuring compliance with accessibility regulations.

One relevant study [28] focuses on assistive technology for cognitive impairment in older individuals, highlighting the importance of compensation systems. These systems employ AI planning techniques to introduce flexibility into tasks like schedule management, personalized reminders, and guidance for routine tasks that improve daily life for these individuals. The system takes on the cognitive load associated with internalizing routine but crucial tasks, reducing the need for external monitoring and instances of compliance failure. Implementing AI in the workplace can enable design principles to assist process users with cognitive disabilities by offering timely reminders, tracking tasks to ensure follow-up, learning routines, and guiding them through tasks.

A UK study [29] explores the use of AI, machine learning (ML), and deep learning (DL) to address regulatory compliance challenges faced by financial institutions. These technologies automate tasks, process complex information, and reduce cognitive load for users. For example, AI-powered voice assistants aid in administrative tasks, scheduling, and accessing information, simplifying work responsibilities. DL algorithms analyze and interpret data, providing insights and recommendations for informed decision-making. These AI technologies empower individuals with cognitive disabilities to overcome challenges related to information processing, memory, and task completion, leading to increased efficiency and effectiveness. In addition, the use of AI, ML, and DL can contribute to compliance with regulations such as Section 508 by facilitating the development of inclusive interfaces and adaptive technologies. Natural language processing capabilities enable voice-controlled interfaces that alleviate the need for complex navigation or manual input, while AI algorithms accommodate different communication styles and preferences, ensuring information accessibility for individuals with cognitive disabilities.

A collection of case studies provide compelling evidence for the positive impact of reducing administrative work on job satisfaction. Research on U.S. physicians [30] revealed that spending less time on administrative tasks is associated with higher career satisfaction. Similarly, a study on general practitioners [31] highlights how excessive paperwork and bureaucratic interference contributed to reduced job satisfaction and increased stress levels. Social workers [32] also reported job dissatisfaction linked to paperwork. Additionally, studies on public sector employees in Switzerland [33] and China [34] emphasized the detrimental effects of red tape on work outcomes, including increased resignation rates, decreased job satisfaction, and heightened procrastination behavior. A study in Australia makes a case for finding ways to support those with dyslexia to prevent job-burnout. The study finds that "excessive job demands, in the absence of supportive job resources and personal resources, leads to poor mental health and wellbeing…results seem to suggest that employees with dyslexia face challenges in the workplace related to their disability including excessive mental exhaustion, and fatigue, leaving them vulnerable to workplace stress and job burnout. Improving psycho-social workplace environments, increasing job resources, decreasing job demands, and critically influencing work engagement, will reduce job burnout and reduce apparent difficulties for individuals with dyslexia in the workplace." [35] By leveraging AI technologies to automate and streamline administrative processes, individuals with cognitive disabilities can experience a significant reduction in paperwork, bureaucratic complexities, and associated stress. This can lead to improved job satisfaction, allowing them to focus more on their core tasks and enhance their overall well-being.

The findings from a South Korean study [36] highlight the importance of individual-level factors, such as job satisfaction and organizational citizenship behavior, in predicting organizational performance in both the United States and Korea. This suggests that addressing these individual-level factors can lead to improved organizational performance. Job satisfaction has been shown to positively influence performance outcomes. By improving job satisfaction, organizations can enhance employee motivation, commitment, and overall engagement, resulting in improved performance. For individuals with cognitive disabilities, AI can play a crucial role in improving job satisfaction by accommodating their specific needs and providing support tailored to their abilities. AI solutions can be developed to streamline administrative tasks, reduce barriers, and enhance accessibility, thereby creating a more inclusive and accommodating work environment. This, in turn, can improve job satisfaction for individuals with cognitive disabilities, leading to enhanced performance for both individuals and organizations.

Organizations like Microsoft are actively working to improve the workplace for individuals with cognitive disabilities, exemplifying the significance of applying AI to business process management. Microsoft's collaboration with Clover Technologies, Concurrency, and Gigi's Playhouse Down syndrome development centers resulted in a mixed-reality platform and Azure-based solution that enables individuals with cognitive disabilities to engage in meaningful warehouse work. [37] This real-world example demonstrates the potential of AI to transform business operations, broaden employment opportunities, and empower individuals with cognitive disabilities. The progress made by Microsoft and its partners underscores the value and potential of AI to foster inclusivity, increase job satisfaction, and enhance performance for individuals with cognitive disabilities.

Cognitive and learning disabilities encompass a wide range of difficulties in cognitive functions, including learning, communication, reading, writing, math, understanding complex information, learning new skills, coping independently, memory, attention, and specific types of thinking. [14] AI solutions exhibit remarkable capabilities such as perfect recall, extensive knowledge, understanding of process-specific language, access to historical process data, awareness of task statuses, and identification of non-compliant tasks. When comparing human cognition to AI cognition, it is important to note that even individuals without cognitive disabilities experience limitations in attention, short-term and long-term memory, recall speed, and executive function. AI solutions specifically designed to enhance accessibility can alleviate the physical and mental burdens for individuals with cognitive disabilities, and even have a positive impact on those without such disabilities as a second-order effect. This reduction in cognitive load and improved accessibility has the potential to increase job satisfaction and ultimately enhance performance, while ensuring compliance with regulatory standards.

## IV. PROCESSGPT CAPABILITIES

The architecture proposed in [12] rigorously outlines textual capabilities of ProcessGPT and additionally introduces multi-modal capabilities that would leverage voice, imagery, and video sources to augment and automate business processes. Designed as an adaptable AI system and trained using existing organizational processes, ProcessGPT would seamlessly integrate into various industries and business processes. Here, we study ProcessGPT's functions in a variety of applications to better understand the benefits it brings to users with cognitive disabilities.

**Filling Out Forms.** Automatically fills out forms at the request of users.

*Example Process:* Employee onboarding. A form with business jargon confuses a new employee. ProcessGPT, understanding the context, recommends the correct field value to the user.

*Cognitive Support:* Simplifies the process by filling out forms, reducing cognitive load and confusion caused by business jargon.

**Answering Process Questions.** Provides guidance on the next steps in a process.

*Example Process:* Project management. A project management system requires a user to manually coordinate tasks outside of the system but does not provide guidance. ProcessGPT provides clear, step-by-step guidance when needed.

*Cognitive Support:* Provides clear, step-by-step instructions, aiding individuals who may struggle with complex instructions.

**Reminders and Follow-ups.** Reminds users to follow up on requests.

*Example Process:* Sales process. A CRM system does not provide reminders for follow-ups. ProcessGPT reminds a salesperson to follow up with a potential client.

*Cognitive Support:* Provides reminders, assisting individuals who may have difficulties with memory or attention.

**Resource Planning.** Analyzes structured data to perform resource planning.

*Example Process:* Team scheduling. A scheduling system is complex and time-consuming. ProcessGPT analyzes team members' schedules and skills to assign tasks efficiently.

*Cognitive Support:* Automates complex tasks like resource planning, reducing cognitive load and the need for multitasking.

**Email Processing.** Processes emails, reducing non-productive time spent on replying, searching, and organizing emails.

*Example Process:* Customer service. An email system lacks efficient sorting and replying features. ProcessGPT helps a customer service rep respond to customer inquiries quickly and effectively.

*Cognitive Support:* Simplifies email processing, aiding individuals who may struggle with organization and prioritization.

**Voice Transcription.** Transcribes calls and meetings.

*Example Process:* Meeting transcription. A transcription system is slow and inaccurate. ProcessGPT accurately transcribes a team meeting and shares the notes with all participants.

*Cognitive Support:* Provides transcription services, aiding individuals who may have difficulties with auditory processing or note-taking.

**Onboarding Assistance.** Assists with onboarding new employees.

*Example Process:* New employee orientation. An orientation system is confusing for new employees. ProcessGPT answers a new employee's questions about company policies in a clear and understandable way.

*Cognitive Support:* Provides clear and understandable answers to questions, aiding individuals who may struggle with complex information.

**Task Management.** Helps manage tasks, extract data, highlight specific items of interest, and review legal agreements.

*Example Process:* Contract management. A contract management system is complex and hard to navigate. ProcessGPT extracts key terms from a contract for review in a user-friendly format.

*Cognitive Support:* Simplifies task management, aiding individuals who may struggle with organization and prioritization.

**Data Interpretation.** Interprets user data, answers queries, and expedites project progress.

*Example Process:* Data analysis. A data analysis system is complex and requires advanced skills. ProcessGPT interprets sales data and provides a forecast for the next quarter in a simple, understandable format.

*Cognitive Support:* Presents data in a simple, understandable format, aiding individuals who may struggle with complex data or numerical thinking.

**Automating Repetitive Tasks.** Automates repetitive and generic tasks.

*Example Process:* Travel claims. A travel system requires manual data entry. ProcessGPT automatically associates credit card charges with travel expenses, creates expenses from receipt images, and enables conversational creation of vouchers, saving time and reducing errors.

*Cognitive Support:* Automates repetitive tasks, reducing cognitive load and the potential for errors.

**User-Centric Guidance.** Provides straightforward, comprehensible guidance and process support tailored to the cognitive capabilities and needs of each individual process user.

*Example Process:* Onboarding. An organization's onboarding process is unclear because it consists of multiple independent sub-processes. ProcessGPT provides a user with step-by-step guidance that bridges sub-processes.

*Cognitive Support:* Provides clear, step-by-step guidance, aiding individuals with limited executive function.

**Intermediary for Internal Systems.** Acts as an intermediary between users and internal systems.

*Example Process:* Timekeeping system. A timekeeping system has substantial latency. ProcessGPT acts as a responsive front end where the user can input their working hours without delays. ProcessGPT then interfaces with the organization's timekeeping system in the background.

*Cognitive Support:* Acts as an intermediary for complex systems, aiding individuals who have limited attention.

**Drafting Email Responses.** Drafts email responses based on user's needs.

*Example Process:* Email communication. A user needs to follow up with a colleague. ProcessGPT drafts an email for the user based on the context of the follow-up.

*Cognitive Support:* Reduces cognitive load by drafting emails, aiding individuals who may struggle with written communication.

**Searching Multiple Data Sources.** Searches multiple data sources based on conversational guidance from the user.

*Example Process:* Data retrieval. A user needs to find a document but can only recall general information about it. ProcessGPT searches through emails, enterprise storage, and local storage based on conversational input to find the document.

*Cognitive Support:* Simplifies the process of searching through multiple data sources, aiding individuals with limited memory.

**Compiling Data from Disparate Sources.** Automatically compiles data from disparate data sources without the user having to search and open each one.

*Example Process:* Data compilation. A user needs to compile data from different sources for a report. ProcessGPT automatically compiles the necessary data, saving the user time and effort.

*Cognitive Support:* Automates the task of compiling data from different sources, reducing cognitive load and the potential for errors.

**ProcessGPT Self-Assessment.** Automatic evaluation of user experience and response quality based on conversation content.

*Example Process:* Procurement. A user needs a procurement form drafted to buy a piece of hardware for a project. ProcessGPT automatically drafts the form but requires correction.

*Cognitive Support:* Enables seamless testing with real users by identifying and implementing corrective feedback. This eliminates typical burdens created by process administrators such as surveys and suggestion boxes.

While this is not an exhaustive list, the table below demonstrates a surjective mapping of ProcessGPT capabilities to the key areas established by the W3C to make content usable for individuals with cognitive disabilities. This demonstrates that ProcessGPT enables knowledge-intensive business processes to accommodate a broad range of cognitive disabilities without altering the underlying systems.

TABLE 1
CORRELATION OF PROCESSGPT FUNCTIONS WITH COGNITIVE DISABILITY ACCOMMODATIONS

| | Accommodation | ProcessGPT Functions Enabling the Accommodation |
|---|---|---|
| 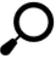 | Help users understand what things are and how to use them | Filling Out Forms, Answering Process Questions, Onboarding Assistance, User-Centric Guidance |
| 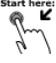 | Help users find what they need | Searching Multiple Data Sources, Compiling Data from Disparate Sources |
| 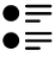 | Use clear content (text, images and media) | Intermediary for Internal Systems |
| 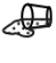 | Help users avoid mistakes | Reminders and Follow-ups, Task Management, Automating Repetitive Tasks |
| 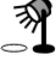 | Help users focus | Email Processing, Drafting Email Responses, Resource Planning, Data Interpretation |
| 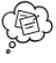 | Ensure processes do not rely on memory | Reminders and Follow-ups, Searching Multiple Data Sources, Voice Transcription |
| 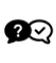 | Provide help and support | Answering Process Questions, Onboarding Assistance, User-Centric Guidance |
| 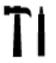 | Support adaptation and personalization | User-Centric Guidance, Intermediary for Internal Systems |
| 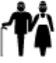 | Test with real users | ProcessGPT Self-Assessment |

## V. QUANTIFYING BENEFITS

Accommodating individuals with cognitive disabilities in knowledge-intensive business processes directly affects process usability, which is a quality attribute associated with the user interface of a system. [38] Business process usability is captured by quality attributes that fall into four categories: quality of function, quality of input & output objects, quality of non-human resources, and quality of human resources. Each category has numerous dimensions, as defined in [39] and shown in the table below.

TABLE 2
QUALITY DIMENSIONS OF BUSINESS PROCESSES

| Function | Input/Output | Non-Human Resource | Human Resource |
|---|---|---|---|
| Suitability | Accuracy | Suitability | Domain Knowledge |
| Accuracy | Objectivity | Accuracy | Qualification |
| Security | Believability | Security | Certification |
| Reliability | Reputation | Reliability | Experience |
| Understandability | Accessibility | Time | Time Management |
| Learnability | Security | Efficiency | Communication Skills |
| Time Efficiency | Relevancy | Resource Utilization | |
| Resource Utilization | Value-added | Effectiveness | |
| Effectiveness | Timeliness | Safety | |
| Productivity | Completeness | User Satisfaction | |
| Safety | Amount of Data | Robustness | |
| User Satisfaction | | Availability | |
| Robustness | | | |

Applying ProcessGPT to a knowledge-intensive business process will improve quality dimensions within the Function and Input/Output categories. Similarly, it reduces the performance requirements of human and non-human resources while delivering the same or better quality in those areas. The improved function and input/output quality provided by ProcessGPT translate directly to a better user experience. Likewise, reducing the performance requirements of non-human and human resources opens up the same user experience to a broader range of users. This collectively demonstrates that ProcessGPT improves usability for all process users–not just those with cognitive disabilities.

The benefits of improving business process usability are well characterized and overwhelmingly positive. Cost savings from users spending less time on a process are most evident; however, second-order savings accrue in areas such as development, support, training, documentation, and maintenance. [40][41] Conventional usability improvements have a near-term cost-benefit ratio ranging from 1:2 for low-volume processes to more than 1:100 for high volume processes. [42][43][44] Usability savings accrue continually such that cost-benefit ratio only improves over time. It is important to note the converse: the costs of poor usability accrue continuously. This is true regardless of whether usability issues are known by process administrators. Ultimately, the users' cognitive factors serve as the testing grounds for process usability.

ProcessGPT can also relieve the administrative burden put onto employees by external organizations. For example, government-mandated forms and processes often have poor usability, which impacts the users and the government (and taxpayers). Government contracts are particularly egregious, with up to 9% of a contract's value spent on administering the contract itself. That means for every dollar spent on services, an organization is spending nine cents on non-value-added activity. [45] Implementing an AI system like ProcessGPT has the potential to significantly reduce the time spent on these non-value-added tasks. Augmenting existing functions such as prioritizing inboxes and automatically drafting basic emails can also have large impacts. Using email accounts for 28% of an employee's time at work. [46] Reducing the time spent searching past emails or documents gives employees more time to focus on the tasks at hand.

Gains from usability improvements are so broad that they are difficult to characterize entirely, yet their impact is undeniable and impossible to overlook. [47] Numerous methods exist to quantify usability benefits, and there is no shortage of relevant case studies. [40][41] It is common to see process time and costs reduced by half as well as absolute savings in the millions of dollars. Usability improvements sometimes pay for themselves within the first day of implementation. [41] In fact, the authors found no research that showed a negative return on usability investments.

Improved usability of business processes also leads to increased productivity and associated cost savings. The same work can be done by less people, or the same people can do more (and more relevant) work. [43] Usable processes reduce the up-front costs needed for documentation and training. Likewise, continuous cost savings are realized in these same areas because usable processes enable process users to be more self-sufficient.

Empowering users to reach a goal via an intuitive, responsive process pays additional dividends in the form of improved employee morale and increased self-directed work. People naturally avoid tasks they don't like, resulting in procrastination behaviors, poor work quality, and increased risk of turnover. [34] These again impact the bottom line of an organization. [43]

## VI. POTENTIAL CHALLENGES

The use of an AI system to help ease bureaucratic burden within an organization is not without challenges. Anytime government or commercial information systems are storing and collating data, there is concern of data security. An all-knowing AI system could have the potential to create proprietary information or even impact the classification of data. Following guidance on AI development and implementation will be central to mitigating these issues. [48] In addition to data security risk, the access to employee correspondence (e.g. email, chats, personal data) could pose a threat to program integrity if there is no human manager to make ethical/tactical decisions on what should be shared.

Beyond its application for ProcessGPT, there is general apprehension about adopting AI technologies. A Gallup poll from 2018 indicated that most people think AI will destroy more jobs than it creates. [49] There is uncertainty around the future of AI's impact on society that can cause anxiety for many. There is also the question of achievability – when will the technology exist to create a working system like ProcessGPT? Additionally, as AI technology is developed, it is imperative to keep humans' emotions at the center. Following best practices such as those laid out in the HAX Toolkit [50] developed by Microsoft will be crucial to making AI an accepted part of organizations. This kind of "human-centric" design will ensure that people using the technology are confident in the benefits they bring and trust that those benefits outweigh the costs.

## VII. CONCLUSIONS

We drew upon an analysis of a proposed Cognitive BPM architecture, ProcessGPT, to understand the potential role of AI in enhancing workplace inclusivity and efficiency, particularly for individuals with cognitive disabilities. We viewed these capabilities through the lens of human cognitive factors and their limitations to address our initial research questions.

**RQ1:** How can AI-intensive technology be applied to business processes to accommodate individuals with cognitive disabilities?

We studied ProcessGPT's functions in hypothetical applications to better understand how it can accommodate users with cognitive disabilities. The functions consisted of:

- Filling out forms
- Onboarding assistance
- Task management
- Answering process questions
- User-centric guidance
- Reminders and follow-ups
- Email processing
- Resource planning
- ProcessGPT Self-Assessment
- Intermediary for internal systems
- Searching multiple data sources
- Automating repetitive tasks
- Drafting email responses
- Voice transcription
- Compiling data from disparate sources
- Data interpretation

We logically demonstrated that ProcessGPT enables knowledge-intensive business processes to accommodate a broad range of cognitive disabilities by correlating these functions with the W3C guidelines for *Making Content Usable for People with Cognitive and Learning Disabilities*.

**RQ2:** What are the benefits of accommodating individuals with cognitive disabilities in business processes?

We correlated ProcessGPT disability accommodations with cognition-agnostic BPM quality attributes to demonstrate that ProcessGPT improves usability for all process users. Generalizing these usability improvements also enables ProcessGPT benefits to be compared to the well-studied and universally positive impacts of usability engineering. Ultimately, usability of a process is correlated with the cognitive demand it places on its users, and no user has unlimited cognitive resources. ProcessGPT and similar AI-intensive systems would improve usability of knowledge-intensive processes for everyone by accommodating individuals with limited cognitive factors, resulting in increased productivity and employee morale across the workforce.

In this work, we address a critical need that affects a significant portion of the workforce by focusing on accommodating individuals with cognitive disabilities. Simultaneously, we illuminate a novel approach to reduce administrative burdens across the workforce. While users without cognitive disabilities must endure cumbersome knowledge-intensive business processes with little recourse, many organizations are legally obligated to accommodate individuals with disabilities. This is comparable to automatic door openers, which benefit all users despite being installed to fulfill legal requirements to accommodate individuals with physical disabilities. Users without physical disabilities benefit from automatic door openers when their arms are full or they are part of a large group, for instance. If a door has poor usability because it is heavy or poorly positioned, automatic openers rectify these issues for everyone. In these instances, disability accommodations rectify flaws in systems that were not properly designed or validated by the original designer. ProcessGPT would perform this function within knowledge-intensive processes across a wide range of use cases. Organizations that are legally obligated to accommodate individuals with disabilities may achieve compliance by applying ProcessGPT or similar AI-intensive systems to their knowledge-intensive business processes. Additional benefits in the form of reduced costs, increased productivity, and improved morale only strengthen this argument